\documentclass[lettersize,journal]{IEEEtran}
\usepackage{amsmath,amsfonts}
\usepackage{algorithmic}
\usepackage{algorithm}
\usepackage{array}
\usepackage{multirow}
\usepackage[caption=false,font=normalsize,labelfont=sf,textfont=sf]{subfig}
\usepackage{textcomp}
\usepackage{stfloats}
\usepackage{url}
\usepackage{verbatim}
\usepackage{graphicx}
\usepackage{cite}
\usepackage{pifont}
\newcommand{\cmark}{\ding{51}}
\newcommand{\xmark}{\ding{55}}

\begin{document}

\title{Enhancing Diversity and Feasibility: Joint Population Synthesis from Multi-source Data Using Generative Models}

\author{Farbod~Abbasi,
        Zachary~Patterson,
        and~Bilal~Farooq%
\thanks{Farbod Abbasi is a PhD student in Information and Systems Engineering at Concordia University, Montreal, QC, Canada (email: farbod.abbasi@concordia.ca).}%
\thanks{Zachary Patterson is a Professor at the Concordia Institute for Information Systems Engineering, Concordia University, Montreal, QC, Canada (email: Zachary.Patterson@concordia.ca).}%
\thanks{Bilal Farooq is an Associate Professor and Canada Research Chair in Disruptive Transportation Technologies and Services at Toronto Metropolitan University, Toronto, ON, Canada (email: bilal.farooq@torontomu.ca).}%
}



\maketitle

\begin{abstract}
Generating realistic synthetic populations is essential for agent-based models (ABM) in transportation and urban planning. Current methods face two major limitations. First, many rely on a single dataset or follow a sequential data fusion and generation process, which means they fail to capture the complex interplay between features. Second, these approaches struggle with sampling zeros (valid but unobserved attribute combinations) and structural zeros (infeasible combinations due to logical constraints), which reduce the diversity and feasibility of the generated data. This study proposes a novel method to simultaneously integrate and synthesize multi-source datasets using a Wasserstein Generative Adversarial Network (WGAN) with gradient penalty. This joint learning method improves both the diversity and feasibility of synthetic data by defining a regularization term (inverse gradient penalty) for the generator loss function. For the evaluation, we implement a unified evaluation metric for similarity, and place special emphasis on measuring diversity and feasibility through recall, precision, and the F1 score. Results show that the proposed joint approach outperforms the sequential baseline, with recall increasing by 7\% and precision by 15\%. Additionally, the regularization term further improves diversity and feasibility, reflected in a 10\% increase in recall and 1\% in precision. We assess similarity distributions using a five-metric score. The joint approach performs better overall, and reaches a score of 88.1 compared to 84.6 for the sequential method. Since synthetic populations serve as a key input for ABM, this multi-source generative approach has the potential to significantly enhance the accuracy and reliability of ABM. 
\end{abstract}

\begin{IEEEkeywords}
Population synthesis, sampling zeros, structural zeros, generative models,  agent based modelling
\end{IEEEkeywords}

\section{Introduction}
\IEEEPARstart{A}{gent}-based Models (ABMs) simulate the behaviors and interactions of individual agents and offer insights into the dynamics and emergent patterns within complex systems such as transportation systems \cite{ziemke2016towards, w2016multi, miller1998integrated} and urban planning \cite{gonzalez2021agent}. ABMs require representation of the entire population of the geographic region that they seek to simulate. Naturally, acquiring detailed individual level data is challenging both because of high costs and privacy concerns \cite{li2023learning}. To address this challenge, it is necessary to synthesize populations from sample data that capture key sociodemographic and behavioral attributes, in order to recreate a realistic representation of the population.  

The creation of synthetic populations has traditionally relied on techniques such as Iterative Proportional Fitting (IPF) \cite{pritchard2012advances, fienberg2004iterative}, Combinatorial Optimization (CO) \cite{muller2010population}, and Markov chain Monte Carlo (MCMC) simulation \cite{farooq2013simulation}. More recently, deep generative models such as Generative Adversarial Networks (GANs) \cite{goodfellow2014generative} and Variational Auto Encoders (VAE) \cite{kingma2013auto} have gained attention for their ability to capture complex, high dimensional data distributions and produce more realistic synthetic samples. However, both traditional and deep learning based methods often struggle with key challenges. We highlight three key contributions of our work that address major gaps in both traditional and deep generative approaches.

First, In population synthesis, census data and travel surveys are commonly used as primary data sources \cite{farooq2013simulation, castiglione2015activity}. Despite offering detailed demographic information, census datasets are typically available only at the aggregate level. While useful for capturing macro-scale patterns, aggregate data fail to preserve joint distributions between variables, which are essential for generating realistic individual agents in Agent Based Models (ABMs) \cite{hofer2018including}. As a result, population synthesis in ABMs often rely on travel surveys, which provide rich socio-demographic data and detailed daily travel diaries at the disaggregate level \cite{habib2020large}. However, relying on a single disaggregate data source also has its own limitations, as it may lack the full range of attributes that are crucial for agent based modelling. This means synthetic populations based on a single dataset may not fully represent the complexity of real world behaviour \cite{hawkins2023multi}. To address this gap, it is useful to incorporate a secondary dataset that includes additional attributes not typically found in the primary dataset. Integrating heterogeneous data sources enables the synthesis of populations that are both behaviorally and demographically realistic. In this study, we propose a multi source GAN approach that jointly integrates census and travel survey data during the generation process, which enables the model to learn and synthesize coherent, complete synthetic individuals in a unified framework. This generative approach leverages the strengths of each dataset while compensating for their individual limitations. It allows the model to learn high dimensional feature relationships and produce realistic, diverse population samples suitable for agent based modeling.

Second, generative models are prone to challenges such as mode collapse, where only limited variations of the data are produced, as well as issues related to sampling zeros and structural zeros. Sampling zeros are feasible combinations of attributes that exist in the real population but are absent in the training data, while structural zeros are impossible or illogical combinations that should never occur in the real world and must be explicitly avoided during generation. Despite their significance, only limited research in the transportation domain has explicitly addressed these issues \cite{kim2023deep, garrido2020prediction}. However, improving the treatment of sampling and structural zeros is crucial for generating high quality and reliable synthetic populations. To mitigate these issues, our approach introduces a regularization term that encourages the generation of diverse and feasible samples to enhance the robustness of the resulting synthetic population.

Third, unlike image or text data, evaluating the quality of generated tabular data is challenging \cite{chundawat2022universal}. A useful population synthesis method must not only replicate observed distributions but also ensure representativeness, novelty, diversity, realism, and coherence of the generated individuals \cite{eigenschink2023deep}. In transportation research, most generative models have relied on metrics such as the Standardized Root Mean Squared Error (SRMSE) and correlation coefficients to assess the similarity between real and synthetic distributions. However, these metrics are often unbounded in scale, lack consistency and scalability, which makes them inadequate for fully assessing the complexity and quality of synthetic datasets. The absence of a uniform evaluation metric makes it difficult to meaningfully compare the effectiveness of different generative approaches. To address this, we apply a unified evaluation metric that aggregates different similarity measures into a single bounded score. This metric ensures comparability across methods, and provides a clear and interpretable assessment of synthetic data quality.

The rest of the paper is organized as follows. The next section reviews the existing literature on population synthesis which highlights both conventional statistical approaches and emerging deep generative methods, while identifying key limitations in current studies. Following this, the methodology section provides the details of the design of the WGAN with regularization term for multi-source data integration. The data and case study section provides details on the datasets used, including their sources and pre-processing steps. In the evaluation metrics and results section, we show how the proposed approach performs in terms of distributional similarity, feasibility, and diversity, with comparisons across different model configurations. Finally, the paper concludes with a discussion of the findings, and suggestions for future research directions.

\section{Literature Review}
Broadly, the methodologies for population synthesis can be categorized into two main groups: conventional statistical approaches and deep generative models. We briefly outline the current state of each method and highlight the main limitations that motivate further research. 

One of the most widely used conventional approaches in population synthesis is the reweighting method \cite{hermes2012review}. This approach adjusts the weights of individuals in a survey sample so that the aggregated distributions match known population margins, typically derived from census data. Reweighting does not generate new individuals and may struggle to capture complex joint distributions across attributes. Matrix fitting \cite{choupani2016population} involves generating individual or household records by fitting them to known marginal distributions using techniques such as IPF \cite{zhu2014synthetic} or e-Maximum Cross Entropy \cite{guo2007population}. Simulation based approaches have emerged as an alternative to matrix fitting techniques for population synthesis. Markov Chain Monte Carlo (MCMC) methods iteratively sample from conditional distributions of attributes to construct synthetic populations that reflect the joint distribution of the real data \cite{farooq2013simulation, saadi2016hidden}. These traditional methods are limited in their ability to model complex relationships, especially as the number of features increases. This restricts their capacity to effectively capture the underlying joint distribution of attributes. Moreover, they are typically designed for single-source datasets and lack any method to address sampling zeros or structural zeros. These shortcomings reduce the representativeness and realism of the synthetic populations.

More recently, deep generative models have been introduced as a promising alternative for population synthesis. Both GANs \cite{goodfellow2014generative} and VAEs \cite{kingma2013auto} have demonstrated significant potential in generating synthetic data that effectively represent real data. They are capable of capturing the statistical properties and joint distributions of the original dataset. This makes them valuable tools in various research and application domains. 

Table GAN \cite{park2018data} and TGAN \cite{xu2018synthesizing} are two notable GAN based models developed for synthesizing tabular data. Table GAN is based on convolutional neural networks, and TGAN uses LSTM networks with attention. TGAN demonstrates superior performance in capturing complex dependencies between attributes and generating higher quality synthetic tables compared to Table-GAN. Badu Marfo et al. \cite{badu2022composite} propose Composite Travel GANs, designed to generate not only tabular but also sequential mobility data. The results demonstrate that CTGAN statistically outperforms the VAE. DATGAN \cite{lederrey2022datgan} has introduced an innovative approach by integrating expert knowledge into the generative process using a Directed Acyclic Graph (DAG). This structure allows DATGAN to control the generation process more effectively and reduce overfitting to noise. These approaches also rely on a single data source, which limits their ability to capture richer, cross-dataset feature dependencies. Furthermore, they do not explicitly address critical issues such as sampling zeros and structural zeros. As a result, the generated data may lack both diversity and feasibility. Moreover, their evaluation is often limited to similarity metrics, which provide an incomplete picture of synthetic data quality.

Borysov et al. used a VAE to generate synthetic populations and demonstrated its potential in modeling high dimensional data compared to traditional methods \cite{borysov2019generate}. One of the strengths of their work is the discussion on sampling zeros and structural zeros. However, despite addressing these important challenges, their approach does not provide a clear evaluation metric to assess how well these issues are handled in the generated data. Garrido et al \cite{garrido2019prediction} and Kim et al \cite{kim2023deep} propose WGAN and VAE to address the issue of sampling zeros and structural zeros. Kim et al. define a loss function to enhance both the diversity and feasibility of the generated data and ensure that the synthetic samples are not only varied but also realistic and consistent with plausible attribute combinations. However, these discussions are limited to single-source datasets and do not address the additional complexity of diversity and feasibility in the context of multi-source data fusion.

Hawkins et al. \cite{hawkins2023multi} propose a multi source data fusion framework for generating a synthetic population enriched with household expenditure and time use attributes. Their method sequentially integrates three national datasets to construct a unified statistical base. Vo et al. \cite{vo2025novel} propose a cluster-based data fusion framework that simultaneously integrates passively collected mobility (PCM) data and household travel survey (HTS) data to generate synthetic populations with rich sociodemographic detail and high spatial heterogeneity. However, both approaches rely on statistical inference and optimization based methods, and do not employ deep generative models such as GANs for sample generation. To the best of our knowledge, only Lee et al. \cite{lee2025collaborative} propose a Collaborative GAN (CollaGAN) that fuses multiple datasets and explicitly evaluates both diversity and feasibility of the generated outputs. Their framework integrates household travel survey (HTS) data and smart card (SC) data within a GAN architecture to jointly generate a synthetic dataset. Our study extends this idea with a WGAN-based framework that jointly integrates census and travel survey data. It introduces a distinct regularization to address sampling and structural zeros, and a unified metric to evaluate similarity, diversity, and feasibility.

To provide a concise overview of the current methods and how they compare across critical dimensions, Table~\ref{tab:literature_comparison} summarizes key studies in terms of their use of deep generative models, integration strategies, treatment of sampling and structural zeros, and evaluation metrics.

\begin{table}[!t]
\caption{Comparative Summary of Existing Methods and Contributions}
\label{tab:literature_comparison}
\resizebox{\columnwidth}{!}{
\begin{tabular}{lccccc}
\hline
Study & \multicolumn{1}{c}{\shortstack[c]{Deep Generative\\Model}} & 
\multicolumn{1}{c}{\shortstack[c]{Multi-source\\Integration}} & 
\multicolumn{1}{c}{\shortstack[c]{Addressing\\Zeros}} & 
\multicolumn{1}{c}{\shortstack[c]{Handling\\Zeros}} & 
\multicolumn{1}{c}{\shortstack[c]{Unified\\Metric}} \\
\hline
IPF / CO / MCMC & \xmark & \xmark & \xmark & \xmark & \xmark \\[3pt] 
TableGAN         & \cmark & \xmark & \xmark & \xmark & \xmark \\[3pt]
TGAN             & \cmark & \xmark & \xmark & \xmark & \xmark \\[3pt]
CTGAN            & \cmark & \xmark & \xmark & \xmark & \xmark \\[3pt]
DATGAN           & \cmark & \xmark & \xmark & \xmark & \xmark \\[3pt]
Borysov et al. \cite{borysov2019generate}   & \cmark & \xmark & \cmark & \xmark & \xmark \\[3pt]
Garrido et al. \cite{garrido2019prediction}   & \cmark & \xmark & \cmark & \xmark & \xmark \\[3pt]
Kim et al. \cite{kim2023deep}      & \cmark & \xmark & \cmark & \cmark & \xmark \\[3pt]
Hawkins et al. \cite{hawkins2023multi}  & \xmark & \cmark & \xmark & \xmark & \xmark \\[3pt]
Vo et al. \cite{vo2025novel}       & \xmark & \cmark & \xmark & \xmark & \xmark \\[3pt]
Lee et al. \cite{lee2025collaborative}      & \cmark & \cmark & \cmark & \cmark & \xmark \\[3pt]
\textbf{This Study} & \cmark & \cmark & \cmark & \cmark & \cmark \\ \hline
\end{tabular}
}
\end{table}

\section{Methodology}
\begin{figure}[H]
    \centering
    \includegraphics[width=\columnwidth]{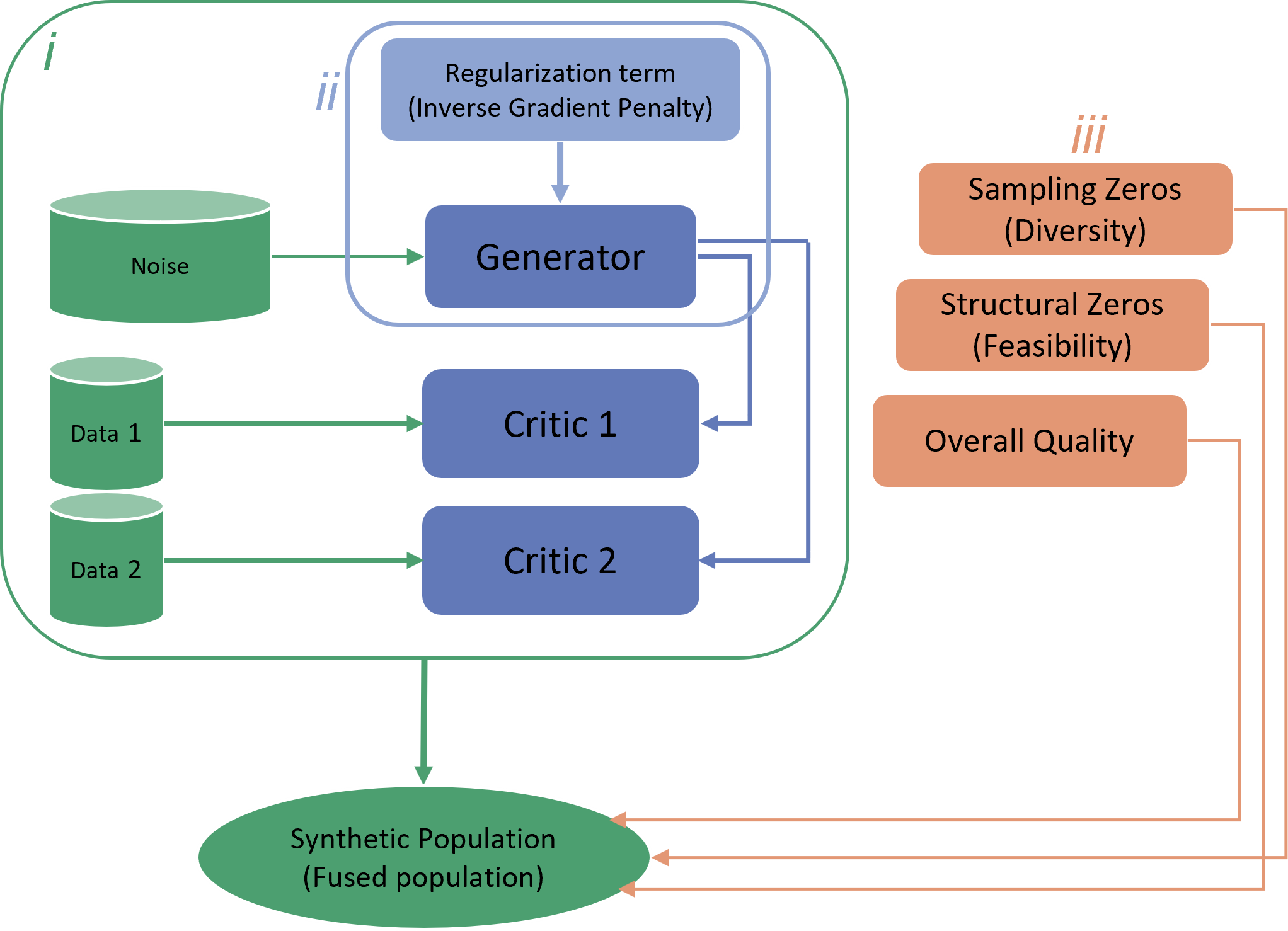}
    \caption{An overall diagram representing the framework}
    \label{fig:framework}
\end{figure}

The proposed framework consists of three main components as shown in Figure~\ref{fig:framework}. \begin{enumerate}
    \item Using GAN to jointly learn from two distinct datasets with common attributes.
    \item Addressing the issue of structural zeros and sampling zeros through the use of a novel loss function.
    \item Using a universal evaluation metric to measure the overall quality of the generated output.
\end{enumerate}

Let \( P_i = \{P_{i1}, P_{i2}, P_{i3}, \dots, P_{im} \} \) denote the set of individuals in the first dataset, where each \( P_i \) represents a unique individual and \( m \) is the total number of features in the first dataset. Similarly, let \( C_j = \{C_{j1}, C_{j2}, C_{j3}, \dots, C_{jn} \} \) represent the set of individuals in the second dataset, where each \( C_j \) corresponds to a unique individual and \( n \) is the total number of features in the second dataset. The resulting synthetic dataset, denoted as \( G_f = \{G_{f1}, G_{f2}, G_{f3}, \dots, G_{fn}, G_{fm} \} \), consists of synthetic individuals, with \( m \) and \( n \) representing the total number of features in the final dataset. It is important to note that both datasets share a subset of common features, which serve as the foundation for integrating the two datasets and generating the final synthetic population. 

\subsection{Generative Adversarial Networks}
Intuitively, GANs comprise two neural networks: a generator and a discriminator, which engage in a mini max game to estimate the joint probability distribution. The generator’s goal is to produce synthetic data that closely resembles real data, whereas the discriminator’s goal is to distinguish between real and synthetic data. During the training process, the discriminator becomes better at distinguishing real samples from synthetic ones, while the generator continually enhances its ability to produce realistic samples. This iterative improvement continues until a Nash equilibrium is reached \cite{nash1950equilibrium}, where both models have optimized their performance to the point where neither can easily outperform the other. The objective function of the GANs is: 
\begin{equation}
\min_G \max_D \mathbb{E}_{x \sim p_{\text{data}}(x)} [\log D(x)] + \mathbb{E}_{z \sim p_z(z)} [\log(1 - D(G(z)))]
\end{equation}
In this equation, \textit{D(x)} represents the discriminator's probability estimate that real data is real, while \textit{D(G(z))} represents the discriminator's probability estimate that synthetic data \textit{G(z)} generated from noise is real. The generator\textit{G} tries to minimize this value to fool the discriminator, and the discriminator \textit{D} tries to maximize it to correctly differentiate between real and fake data. While the traditional GAN has shown remarkable success in generating realistic synthetic data, it faces several challenges, particularly during training. GANs are known to suffer from issues like mode collapse \cite{thanh2020catastrophic}, where the generator produces limited diversity in its outputs, and instability \cite{al2024optimization}, especially when the discriminator is too powerful. These issues often make GANs difficult to train effectively and can lead to suboptimal results. To overcome these challenges, we use the Wasserstein GAN (WGAN) \cite{arjovsky2017wasserstein} with a gradient penalty (GP) \cite{gulrajani2017improved}, a more stable and reliable variant of GAN.  

WGAN with GP minimizes the Wasserstein distance (WD), instead of Jensen Shannon divergence (JSD), which provides a smoother gradient, avoids the vanishing gradient problem, and ensures more stable training. Additionally, WGAN-GP is less prone to mode collapse because the WD directly measures the difference between the real and generated data distributions, which encourage the generator to produce more diverse outputs. The gradient penalty further stabilizes training by regularizing the model and preventing excessively large or small gradients. 

\begin{equation}
\begin{split}
L_D = &\; \mathbb{E}_{x \sim p_{\text{data}}(x)}[D(x)] 
      - \mathbb{E}_{z \sim p_z(z)}[D(G(z))] \\
      &\; + \lambda \mathbb{E}_{\hat{x} \sim p_{\hat{x}}}
      \left[\left(\|\nabla_{\hat{x}} D(\hat{x})\|_2 - 1\right)^2\right]
\end{split}
\label{eq:dis}
\end{equation}

\begin{equation}
L_G = - \mathbb{E}_{z \sim p_z(z)}[D(G(z))]
\label{eq:gen}
\end{equation}
Based on Equation (\ref{eq:dis}) the discriminator loss consists of three terms. The first term, encourages the critic to assign high values to real data. The second term, ensures that fake data receives low values. The third term, is the gradient penalty, that enforces the 1-Lipschitz constraint and stabilizes training by penalizing large gradients. The generator loss, aims to minimize the critic's ability to distinguish fake data from real, and pushes the generator to create more realistic data as shown in Equation (\ref{eq:gen}). 

We aim to use the strengths of GAN to introduce several key contributions to the field of synthetic population generation. First, our framework enables the joint integration and generation of multi-source data instead of using sequential integration and generation or relying on a single dataset. Second, we introduce a regularization term based on inverse gradient penalty that explicitly handles both sampling zeros and structural zeros. This allows the model to generate samples with higher diversity and feasibility. Third, we design a unified evaluation metric that includes diversity, feasibility, and distributional similarity. These indicators result in a more comprehensive and interpretable assessment of synthetic population quality.

\subsection{Joint Generative Adversarial Network}

The proposed framework employs a WGAN-GP to generate a synthetic population using two datasets. Our GAN model consists of three main components: the generator, and two critics, designed to handle different parts of the generated data. The model uses the WGAN-GP framework, which optimizes the Wasserstein distance between the real and generated data distributions and incorporates a gradient penalty to enforce the 1-Lipschitz constraint for stable training. 

\begin{figure}[H]
    \centering
    \includegraphics[width=0.9\linewidth]{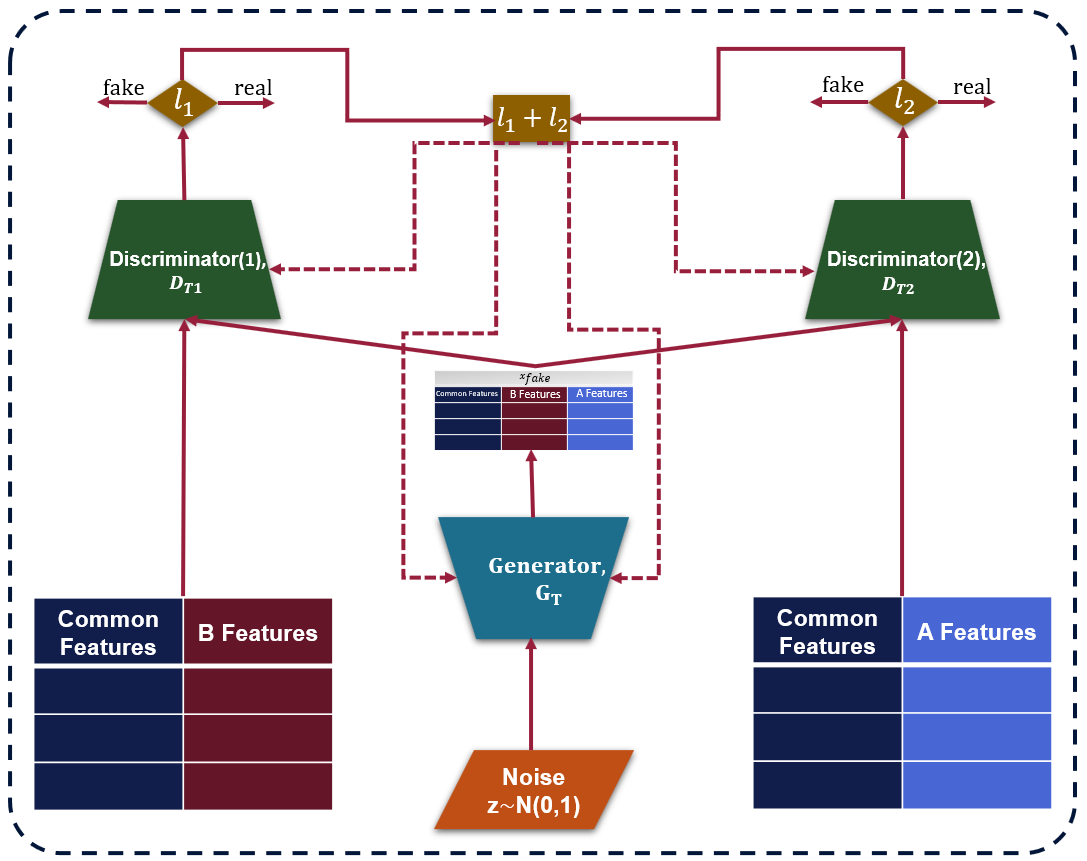}
    \caption{The training procedure of Joint GAN}
    \label{fig:training procedure}
\end{figure}

Based on the Figure ~\ref{fig:training procedure}, the generator network takes random noise as an input and produces synthetic data that includes attributes from both datasets. The model employs two critics to assess the validity of the generated data. Each critic is responsible for evaluating different parts of the generated data. The first critic processes a subset of the generated data corresponding to the first dataset, while the second critic handles another subset from the second dataset. The critics are trained to assign higher values to real data and lower values to generated data. This dual critic design ensures that each part of the generated sample is assessed with specialized focus, which is particularly important when the datasets differ in their feature space or statistical properties. By splitting the evaluation task, each critic can independently guide the generator to better capture the specific distributional characteristics of each dataset. 

In sequential methods, the fusion step typically relies on statistical matching or imputation techniques, which may introduce errors or fail to preserve the true relationships between attributes from different sources. In contrast, our joint generation approach allows the model to directly learn how features from each dataset relate to one another during the training process. This enables the generator to capture more realistic and consistent combinations of attributes, especially when important cross dataset patterns exist. As a result, the generated individuals are more behaviorally and demographically coherent, which makes the synthetic population more representative of real world diversity.

The training process involves alternately updating both the critics and the generator. In each iteration, the critics are trained to distinguish between real and fake data by computing the Wasserstein loss and applying a gradient penalty to enforce the 1-Lipschitz constraint. Once the critics are updated, the generator is trained to minimize the combined loss, which includes the Wasserstein loss from both critics, and an inverse gradient penalty. This promotes output diversity in response to changes in the latent input. 

\begin{table}[H]
\centering
\caption{Model Hyperparameters and Settings}
\label{tab:hyperparams}
\begin{tabular}{p{3.5cm} p{4cm}} 
\hline
\textbf{Items} & \textbf{Values} \\
\hline
Optimizer & \textbf{Adam}, RMSProp \\[3pt]
Generator’s Learning Rate & 0.00005, 0.00002, \textbf{0.0001} \\[3pt]
Critic’s Learning Rate & 0.00005, \textbf{0.00002}, 0.0001 \\[3pt]
Batch Size & \textbf{512}, 1024 \\[3pt]
Epoch & 1001, 2001, \textbf{5001} \\[3pt]
Activation Function & ReLU, \textbf{LeakyReLU} \\[3pt]
Generator Hidden Layer & 5 \\[3pt]
Critic Hidden Layer & 3 \\[3pt]
N\_critic & 3, 4, \textbf{5} \\[3pt]
Gen. Layer Size & \textbf{(18,18,200,100,50)}, (18,18,200,200,100), \newline (16,16,100,100,50) \\[3pt]
Critic Layer Size & \textbf{(256,128,64)}, (128,64,32), \newline (256,256,128) \\[3pt]
Batch Normalization & True \\
\hline
\end{tabular}
\end{table}

To ensure the robustness and stability of the proposed GAN model, we conducted a hyperparameter tuning process. Various configurations were tested for key parameters and the final hyperparameter values, which are highlighted in Table ~\ref{tab:hyperparams}, were selected. These selected settings represent the most effective configuration for our dataset and evaluation goals.

\begin{figure}[H]
    \centering
    \includegraphics[angle=270,width=\columnwidth]{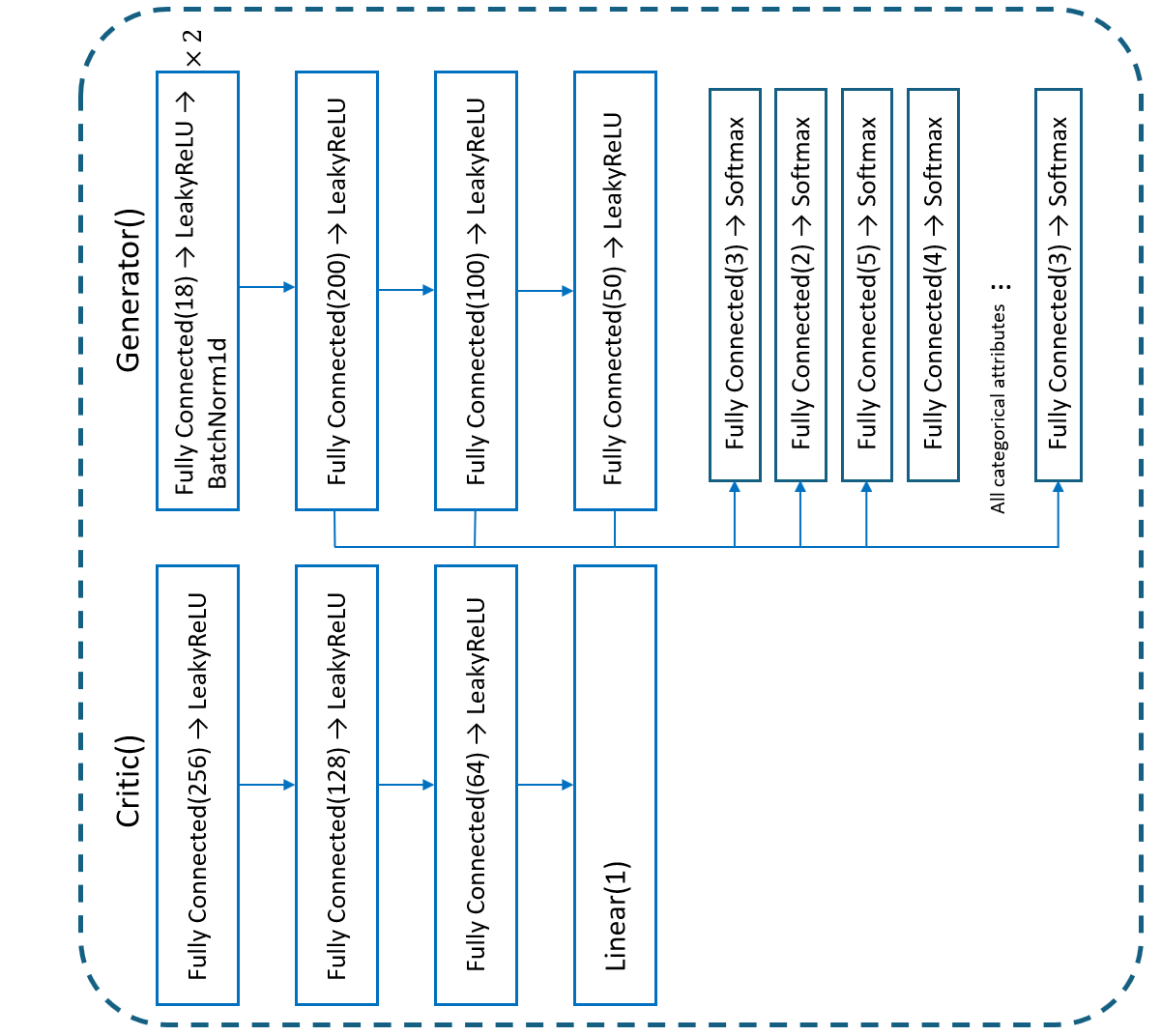}
    \caption{Calibrated architecture and hyperparameters of the WGAN}
    \label{fig:Calibrated}
\end{figure}

Based on Figure \ref{fig:Calibrated}, the generator has two hidden layers. The first layer maps the noise to a space of size 18, using a fully connected linear layer, LeakyReLU activation, and Batch Normalization for stability. The second hidden layer maps the output to a space of size 18, again using LeakyReLU and Batch Normalization. Next, the generator branches into multiple sub networks, each responsible for generating a specific attribute subset. Each branch consists of fully connected layers with LeakyReLU activations and Batch Normalization. These layers progressively reduce the number of units from 200 to 100 to 50, and then the output of each branch is mapped to the desired output size. Each output corresponds to a particular subset of features. A Softmax activation function is applied at the end of each branch to ensure the output values lie between 0 and 1, suitable for categorical data. Both critics use a multi layer fully connected network with LeakyReLU activations, first mapping the input to 256 units and then reducing it to 128 units and 64 units. The output of each critic is a single scalar, that represents the validity of the input data. 

\subsection{Inverse Gradient Penalty}

We should guarantee the generation of diverse and feasible outputs, especially when performing multi-source population synthesis. As the number of features increases, the risk of limited variation in the generated samples becomes more pronounced due to combining datasets with partially common attributes. In such cases, the generator may overfit to dominant patterns or struggle to explore rare but valid feature combinations. To address this, we introduce a regularization term based on the Inverse Gradient Penalty (IGP), which encourages diversity in the generated samples. This method prevents the generator from collapsing to a narrow set of outputs and instead promotes exploration of a wider range of feasible synthetic individuals. As a result, IGP directly enhances the model’s ability to generate diverse and feasible outputs across the joint feature space constructed from multiple data sources. The IGP is introduced to enforce a minimum Lipschitz constant for the generator, and ensures that small changes in the latent space lead to sufficiently large changes in the generated samples \cite{li2023systematic}. This is done by maximizing the expected value of the minimum ratio between the generated output distance and the latent space distance, as shown in Equation (\ref{eq:IGP}).

\begin{equation}
\max_{G} \mathcal{L}_{z}(G)
= \max_{G}\,
\mathbb{E}_{z_{1},z_{2}}
\Bigl[
  \min\!\Bigl(
    \frac{\|\,G(y,z_{1}) - G(y,z_{2})\|}{\|\,z_{1} - z_{2}\|},
    \tau
  \Bigr)
\Bigr]
\label{eq:IGP}
\end{equation}

This regularization term is added to the generator's standard adversarial loss. The generator's primary objective remains to fool the critic, but with the additional constraint that its outputs must change sufficiently when the latent code changes. Based on Equation (\ref{eq:gen_with_reg}), this combined loss ensures that the generator not only learns to produce realistic samples but also maintains diversity across its outputs.

\begin{equation}
L_G = - \mathbb{E}_{z \sim p_z(z)}[D(G(z))] - \lambda \cdot \mathcal{L}_z(G)
\label{eq:gen_with_reg}
\end{equation}

While adding the IGP leads to more diverse outputs, which increase the variety of generated samples, it is important to note that this can also increase the risk of producing infeasible outputs. The key challenge is that diversity in the generated samples might sometimes include infeasible outputs if the model has not fully captured the underlying constraints of the data. Therefore, while the diversity helps reduce mode collapse, careful evaluation is required to ensure that the generated samples are still realistic and feasible. To assess the reliability of the synthetic samples, we use metrics such as recall, precision, and the \textit{F1} score. These metrics provide insight into how well the generated samples avoid structural zeros and sampling zeros issue. 

To calculate recall in \cite{kim2023deep}, the focus is on measuring the proportion of the entire generated sample that overlaps with the full population distribution. While this approach captures overall similarity, it may overestimate the model's generalization ability, since  many generated samples are already similar to the training data. However, the goal of recall in the context of sampling zeros is to assess how many unique and previously unobserved feature combinations the model can generate that exist in the real population but were missing from the training data.

\begin{figure}[H]
    \centering
    \includegraphics[width=1\linewidth]{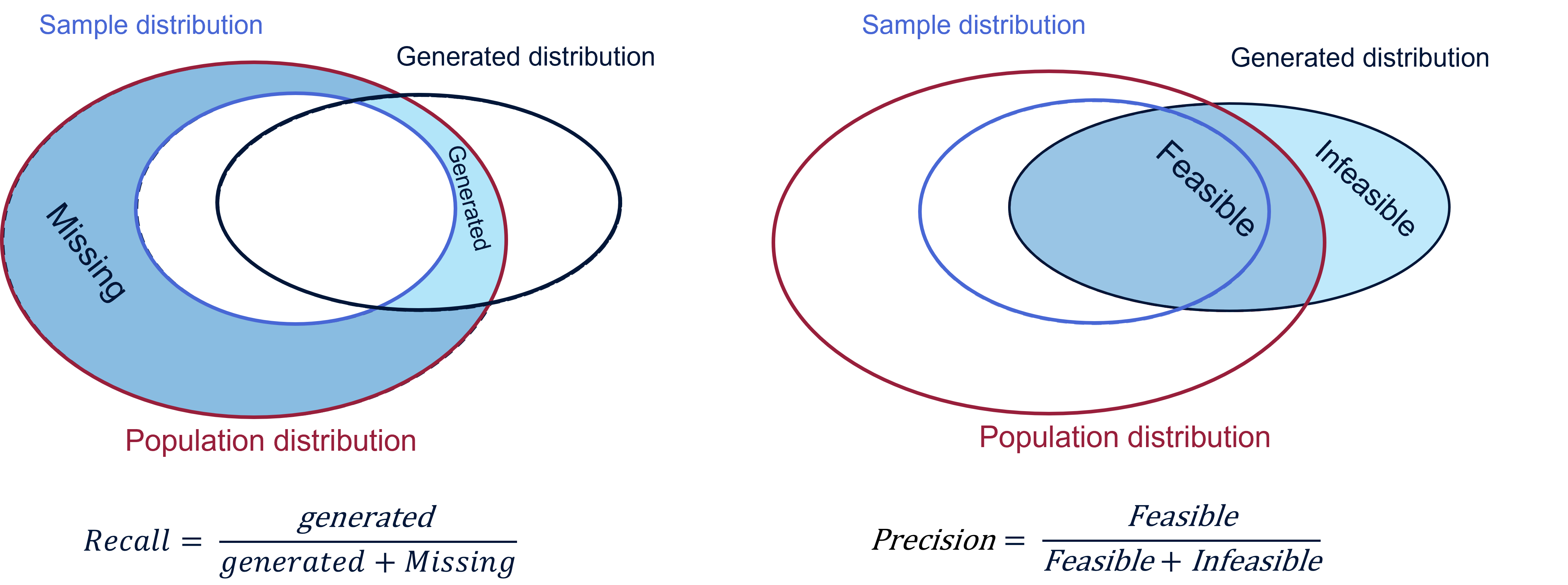}
    \caption{Conceptual diagram of precision and recall based on sampling zeros and structural zeros }
    \label{fig:condiag}
\end{figure}

We suggest a different approach to enhance generalization. As shown in Figure~\ref{fig:condiag}, instead of simply comparing the generated samples with the overall population distribution, it would be more insightful to measure the portion of generated samples that fall outside the training sample but within the population distribution. This would provide a more robust measure of generalization by assessing how well the generator can generalize to unseen or out of sample regions of the population distribution.

\subsection{Evaluation Metrics}
Most transportation research typically relies on simplified distributional similarity and similar metrics to assess the quality of generated datasets\cite{kim2023deep}. However, these metrics risk overfitting to marginal and bivariate distributions, which do not fully capture the correlations in the data. As a result, models evaluated using such metrics may seem to perform well, even if they fail to reproduce higher order dependencies or generate realistic variability. For example, a model that merely replicates observed samples can achieve low SRMSE and high correlation scores, yet still lack true diversity or feasibility. Therefore, it is important to expand our evaluation metrics and make them better. 

\textit{1) Statistical Distance Measure:} We measure \textit{SRMSE} and \textit{JSD} for marginal, bivariate, and trivariate distributions to assess the quality of the synthetic dataset at different levels of complexity. Based on Equation (\ref{eq:srmse}), The \textit{SRMSE} measures the standard root mean squared error between the real and synthetic datasets, and the \textit{JSD} quantifies the divergence between the probability distributions of the real and synthetic datasets as shown in Equation (\ref{eq:jsd}). Finally, $S_{\text{distance}}$ combines the \textit{SRMSE} and \textit{JSD} scores, that gives a unified metric to evaluate the statistical similarity between the real and synthetic datasets across these different distribution levels. The resulting value ranges from 0 to 1, with higher values indicating better similarity. 

\begin{equation}
\begin{split}
\text{SRMSE}(\pi, \hat{\pi}) 
&= \frac{\text{RMSE}(\pi, \hat{\pi})}{\overline{\pi}} \\
&= \sqrt{
  \frac{
    \sum_{(k,k')} \bigl(\pi_{(k,k')} - \hat{\pi}_{(k,k')}\bigr)^2 / N_b
  }{
    \sum_{(k,k')} \pi_{(k,k')} / N_b
  }
}
\end{split}
\label{eq:srmse}
\end{equation}

\begin{equation}
\begin{split}
\mathrm{JSD}_{\pi_1,\ldots,\pi_n}(P_1,P_2,\ldots,P_n)
&= \sum_{i=1}^n \pi_i \, D\bigl(P_i \,\|\, M\bigr) \\
&= H(M) - \sum_{i=1}^n \pi_i \, H(P_i)
\end{split}
\label{eq:jsd}
\end{equation}

\begin{equation}
S_{\text{distance}} =1- \frac{1}{2} 
\left( \sum_{i=1}^{3} \text{SRMSE}_i + \sum_{i=1}^{3} \text{JSD}_i \right),
\label{eq:sdistance}
\end{equation}

\textit{2) Correlation:}
We measure the similarity between the correlation matrices of the real and synthetic datasets as shown in Equation (\ref{eq:scorr}). It compares the first, second and third correlations between the features in the real dataset, \( r_{ij} \), and the corresponding correlations in the synthetic dataset, \( f_{ij} \). The formula computes the logarithmic transformed relative error between these correlations. A higher value of \( S_{corr} \) indicates that the synthetic data better preserves the correlation structure of the real data, with values closer to 0 representing low similarity. This metric is particularly useful for assessing how well the relationships between features in the synthetic data match those in the real data.

\begin{equation}
S_{\text{corr}} = 1 - \frac{1}{n^2 - n} \sum_{i=1}^{n} \sum_{\substack{j=1 \\ j \neq i}}^{n} 
\left| \frac{\text{ln}(r_{ij}) - \text{ln}(f_{ij})}{\text{ln}(r_{ij})} \right|,
\label{eq:scorr}
\end{equation}

\textit{3) $S_{\text{pmse}}$ Index:}
We evaluate the prediction error of a logistic regression model trained on a combined dataset consisting of both real and synthetic samples. The label \textit{Y} is used to distinguish real data (labeled 0) from synthetic data (labeled 1). The model predicts the probabilities $\hat{p}_j$ for each sample, which represents the likelihood of being synthetic. The constant \textit{c} is the ratio of synthetic data samples to the total number of samples. Based on Equation (\ref{eq:spmse_double_sum}), $S_{\text{pmse}}$ quantifies the mean squared error between the predicted probabilities and the true class labels to have an assessment of how well the synthetic data approximates the input data's characteristics. Since the constant \textit{c} is 0.5, the ideal scenario occurs when the predicted probabilities are centered around 0.5, which indicates that the model cannot distinguish between real and synthetic samples.  

\begin{equation}
S_{\text{pmse}} = \frac{1}{N \cdot N} \sum_{i=1}^{N} \sum_{j=1}^{N} \left( \hat{p}_j - c \right)^2,
\label{eq:spmse_double_sum}
\end{equation}

\begin{equation}
c = \frac{N_{\text{syn}}}{N_{\text{real}} + N_{\text{syn}}},
\label{eq:c_value}
\end{equation}

\textit{4) Similsrity Distribution \( S_{\text{cr}} \):}
We measure the coverage of synthetic data points relative to real data. Where \( n_{\text{cat}} \) is the total number of categories, \( n_i^f \) is  the number of synthetic data points, and \( n_i^r \) is the number of real data points for each category. The scaling factor is the ratio of the number of samples in real data and the number of samples in fake data. 

\begin{equation}
S_{\text{cr}} = \frac{1}{n_{\text{cat}}} \sum_{i=1}^{n_{\text{cat}}} \frac{n_i^f}{n_i^r} \times \text{scaling factor},
\label{eq:scr}
\end{equation}

\textit{5) Machine Learning Efficacy \( S_{\text{ml}} \):}
This metric assesses whether synthetic data can serve as a reliable substitute for real data in downstream tasks \cite{bourou2021review}. Specifically, it involves training a model on the generated dataset and testing it on the real dataset to evaluate how well the synthetic data captures meaningful patterns and supports generalization. We train several models such as Random Forest, AdaBoost, Gradient Boosting, and Logistic Regression, on both datasets and calculate the relative errors for each model. A higher value of \( S_{\text{ml}} \) indicates that the synthetic dataset effectively supports machine learning tasks, closely resembling the real dataset in terms of predictive performance.

\begin{equation}
S_{\text{ml}} = 1 - \frac{e_{m1} + e_{m2} + e_{m3} + e_{m4}}{4},
\label{eq:sml}
\end{equation}

\section{Data and Case Study}

\begin{table*}[!t]
\centering
\caption{Synthetic Population Data with Proportions}
\label{tab:variables}
\resizebox{\textwidth}{!}{%
\begin{tabular}{l l c l c}
\hline
\textbf{Attribute (Dimensions)} & \textbf{Category} & \textbf{Proportion (\%)} &
\textbf{Category} & \textbf{Proportion (\%)} \\
\hline
\multirow{1}{*}{Gender (2)} & Male & 48.84 & Female & 51.15 \\[4pt]

\multirow{3}{*}{Driver’s license (5)} & Yes & 70.68 & No & 12.82 \\
& Refusal & 0.01 & Not applicable (under 16 years) & 16.44 \\
& Does not know & 0.004 & - & - \\[4pt]

\multirow{2}{*}{Age (3)} & Under 14 years & 15.35 & 15--60 years & 63.57 \\
& More than 60 years & 21.07 & - & - \\[4pt]

\multirow{3}{*}{Number of households (5)} & 1 & 12.55 & 4 & 23.59 \\
& 2 & 32.86 & 5 & 13.29 \\
& 3 & 17.71 & - & - \\[4pt]

\multirow{2}{*}{Employment Status (4)} & Full time & 40.40 & Not in labour force & 49.93 \\
& Part time & 4.60 & Unemployed & 5.07 \\[4pt]

\multirow{2}{*}{Mobility (3)} & Yes & 79.10 & No & 17.02 \\
& Does not know & 3.87 & - & - \\[4pt]

\multirow{3}{*}{Number of trips (6)} & 1 & 22.95 & 4 & 4.65 \\
& 2 & 53.67 & 5 & 1.01 \\
& 3 & 17.54 & 6 or more & 0.20 \\[4pt]

\multirow{4}{*}{Number of vehicles (7)} & 0 & 9.92 & 4 & 2.95 \\
& 1 & 36.65 & 5 & 0.75 \\
& 2 & 39.94 & 6 or more & 0.42 \\
& 3 & 9.82 & - & - \\[4pt]

\multirow{3}{*}{Number of modes (5)} & 1 & 95.18 & 4 & 0.0008 \\
& 2 & 4.03 & 5 or more & 0.0001 \\
& 3 & 0.07 & - & - \\[4pt]

\multirow{2}{*}{Home type (4)} & Apartment & 55.69 & Single house & 36.15 \\
& Other & 7.88 & Movable & 0.29 \\[4pt]

\multirow{2}{*}{Education level (3)} & Post secondary & 51.18 & Secondary & 26.88 \\
& No certificate & 21.94 & - & - \\[4pt]

\multirow{1}{*}{Tenure (2)} & Owner & 58.21 & Renter & 41.79 \\[4pt]

\multirow{2}{*}{Income level (4)} & Level 1 & 55.77 & Level 2 & 25.95 \\
& Level 3 & 11.28 & Level 4 & 6.99 \\[4pt]

\multirow{1}{*}{Marital status (2)} & Married & 45.96 & Single & 54.04 \\
\hline
\end{tabular}%
}
\end{table*}

The analysis uses two datasets covering the same geographic region: travel data from the 2018 Montreal OD survey and demographic data from the 2016 census. The OD dataset encompasses detailed travel information for 153,399 individuals. The first dataset captures various attributes at the individual, household, and trip levels. The second dataset includes several common features with the OD survey, such as age, sex, employment status, and household characteristics, while also providing additional attributes that are not present in the OD dataset but aggregated to the level of census tract. By combining these two datasets, the study aims to create a robust and enriched database that captures a wide range of demographic and travel behavior characteristics. A list of the variables synthesized in the population is shown in Table~\ref{tab:variables}.

\subsection{Data pre-processing} 

Given that the OD and census datasets share several common attributes, it was essential to ensure consistency in their categorical values distribution. This step involved aligning the categorical values of shared features, such as age, gender, employment status, and household characteristics, to ensure they have the same distribution across both datasets. 

To ensure consistency and comparability across the datasets, normalization was performed on categorical features. Categorical features were normalized using one-hot encoding. This technique converts categorical variables into a series of binary variables, each representing a unique category with values of 0 or 1.

Since the OD dataset is at the disaggregate level, it was crucial to convert the census data into a disaggregate level as well. This conversion is critical for ABMs, which require individual level data. To achieve this, we leveraged the power of GAN to generate synthetic individual level datasets that align with the statistical properties of the aggregate census data. To achieve this, we generated synthetic individuals that preserve the statistical characteristics of the original aggregate census data, while ensuring consistency between the disaggregated and aggregated distributions.  

\section{Results and Discussion}

To demonstrate the robustness of the proposed methodology, three different models were employed. In the first model, instead of simultaneously integrating and generating data from multiple sources, the datasets were integrated separately, and the synthetic data was generated afterward. In the second model, we implemented WGAN-GP without any regularization terms. The third and best model simultaneously integrated and generated data from multiple sources, with an added regularization term. 

One of the key advantages of this evaluation is that it extends the analysis beyond simple similarity in statistical distributions by providing a more comprehensive and quantitative evaluation. In this study, we employ unified metrics to evaluate the quality of the generated dataset \cite{chundawat2022universal}. Each metric is scaled between 0 and 1 and contributes equally to the final score. The metrics include statistical distance measures, correlation, $S_{\text{pmse}}$ index, support coverage, and machine learning efficacy. Additionally, to evaluate the feasibility and diversity of the generated data, we use precision, recall, and F1-score. Recall corresponds to sampling zeros (missing data points in the synthetic data), while precision addresses structural zeros (missing relationships between attributes) \cite{kim2023deep}. The F1-score then combines recall and precision, which is as a final evaluation metric based on the trade off between sampling zeros and structural zeros. These metrics provided a detailed evaluation of the model’s ability to generate realistic, diverse individual level data while maintaining consistency with the original datasets’ statistical properties. 

\subsection{Evaluation Score}

We evaluate different aspects of similarity between the real and synthetic data by calculating five distinct metrics. As shown in Equation (\ref{eq:tabsyndex}), these metrics are then combined to form a unified evaluation score. This composite score provides a comprehensive measure of the overall quality of the GAN models. In the Table \ref{tab:model_comparison}, we present the detailed results for each individual metric.

\begin{equation}
\text{Score} = \frac{S_{\text{distance}} + S_{\text{corr}} + S_{\text{pmse}} + S_{\text{cr}} + S_{\text{ml}}}{5},
\label{eq:tabsyndex}
\end{equation}

 The first component assesses distributional similarity using the SRMSE and JSD across marginal, bivariate, and trivariate distributions. The Joint GAN with regularization consistently achieves lower SRMSE and JSD values than the other models. As shown in Table~\ref{tab:model_comparison}, the SRMSE and JSD increase with distribution complexity from marginal to trivariate, but the increase is consistently less steep for the first model and maintains the lowest SRMSE and JSD values across all distribution types.

\begin{figure}[!ht]
    \centering
    \includegraphics[width=\columnwidth]{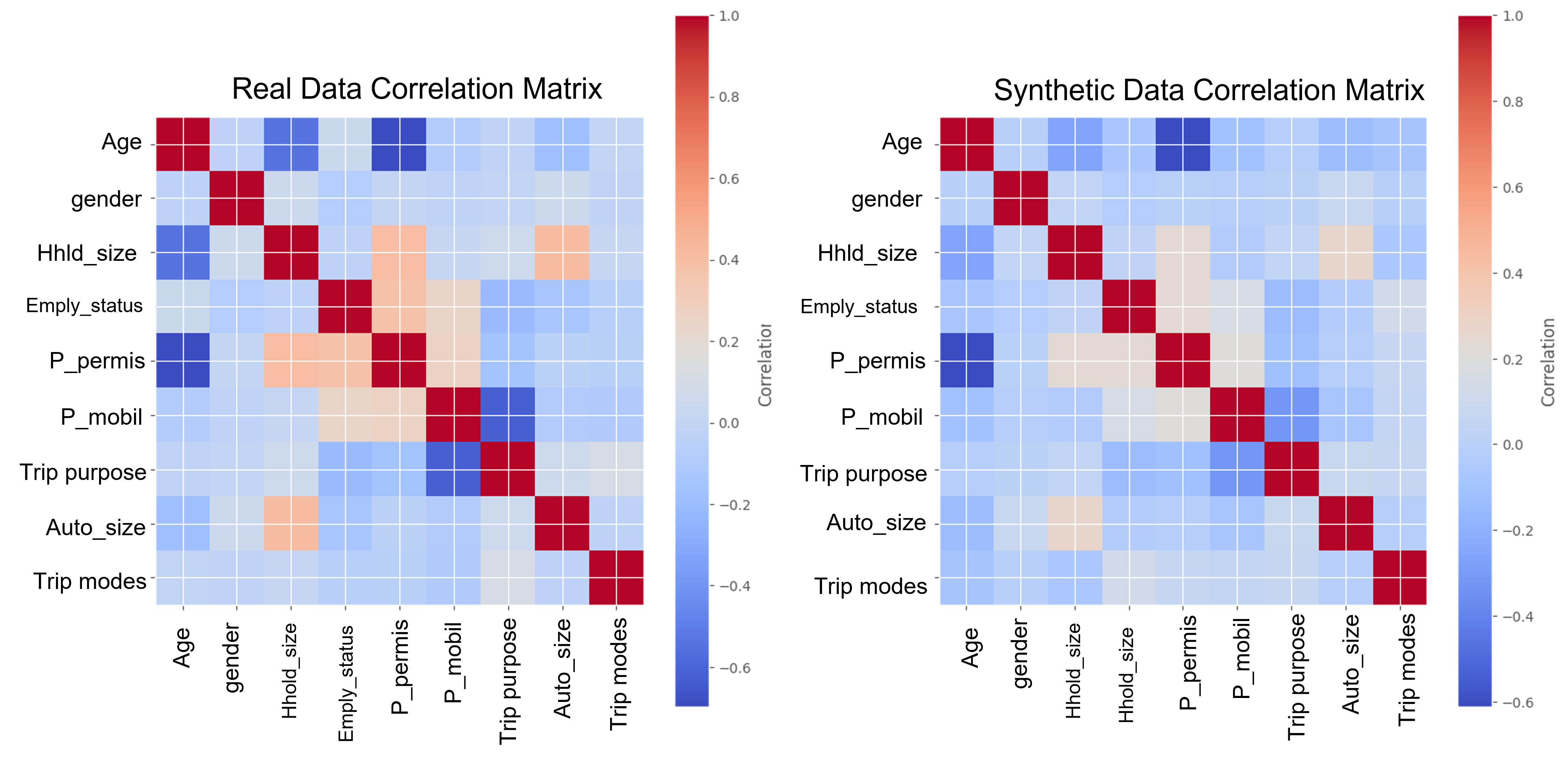}
    \caption{Correlation between the columns of real and synthetic datasets}
    \label{fig:Correlation}
\end{figure}

 The second metric evaluates how well the synthetic data maintains the relationships between features. We measure first, second, and third order correlations, and similar to the previous metric, the results show that the Joint GAN with regularization outperforms the other models. Based on Table~\ref{tab:model_comparison}, when the regularization term is added, the first and third order correlation errors decrease by approximately 22\% and 21\% respectively, compared to the version without regularization. In contrast, there is no significant improvement observed between the Simple GAN and the Joint GAN without regularization. Figure \ref{fig:Correlation} illustrates the correlation structures of the real and synthetic datasets. The synthetic dataset preserves the overall correlation patterns observed in the real data. This suggest that the generative model effectively captures the underlying relationships among features.

\begin{figure}[!ht]
    \centering
    \includegraphics[width=1\columnwidth]{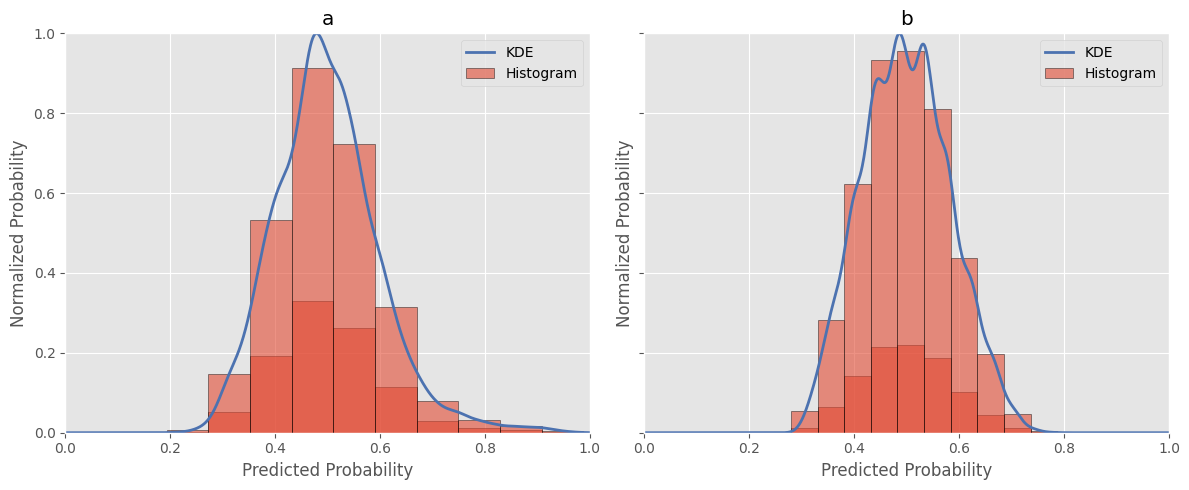} 
    \caption{\scriptsize (a) Likelihood of Being Synthetic for OD dataset, (b) Likelihood of Being Synthetic for Census dataset}
    \label{fig:PMSE}
\end{figure}

The third metric measures how well the synthetic dataset is indistinguishable from real data. Based on Table ~\ref{tab:model_comparison} the Joint GAN with regularization achieves the average lowest PMSE value of 0.144 Among the models. As shown in Figure \ref{fig:PMSE}, the distribution of prediction probabilities for OD and census datasets shows the peaks near 0.5. This suggests that the synthetic dataset closely resembles the real data.

\begin{figure}[!ht]
    \centering
    \includegraphics[width=1\columnwidth]{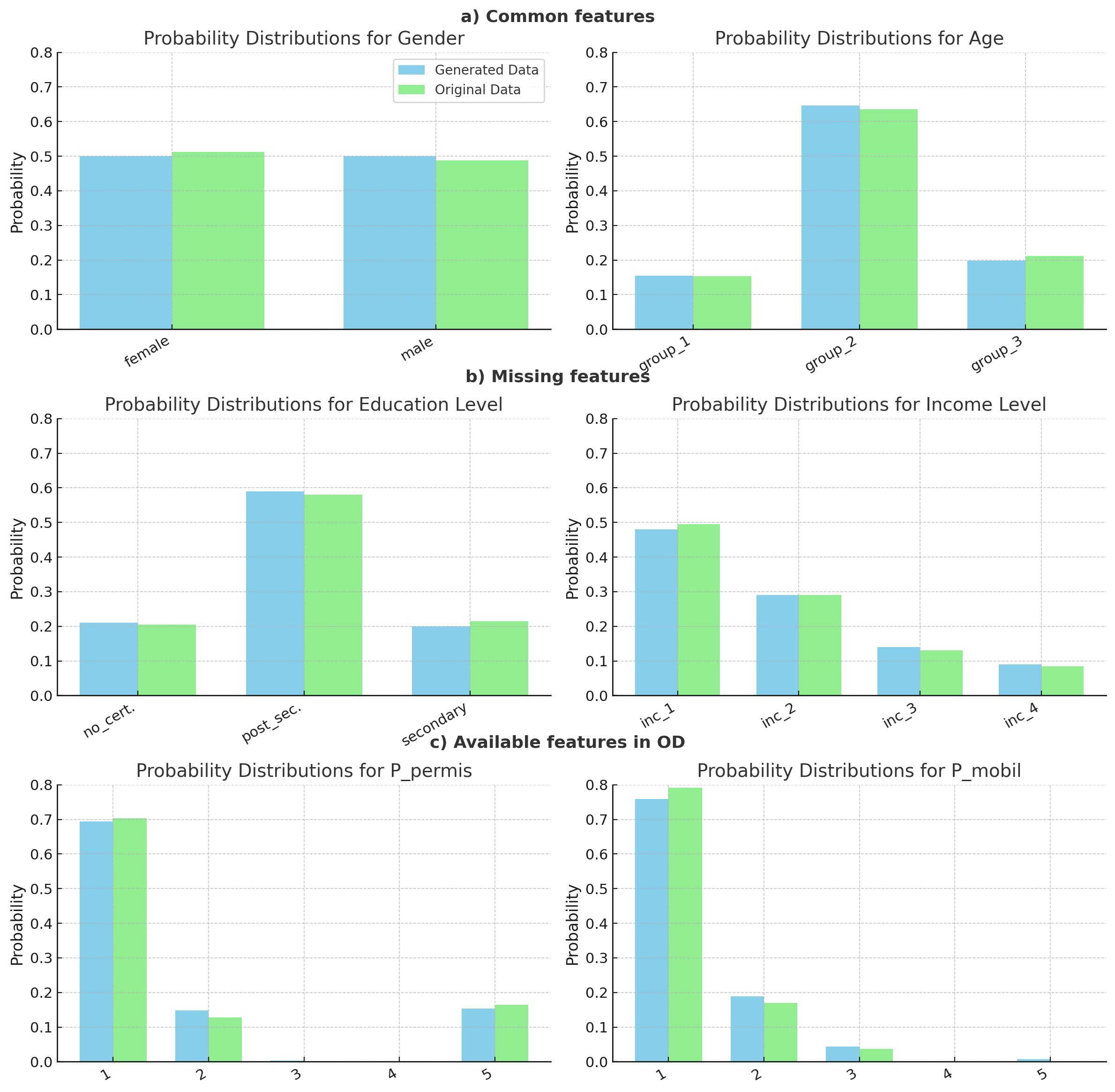} 
    \caption{\scriptsize Probability distribution of input and generated datasets for (a) common features, (b) missing features, (c) Available features }
    \label{fig:similarity}
\end{figure}

The fourth metric captures the coverage of synthetic data compared to real data categories. Figure \ref{fig:similarity} depicts the distributional similarity of the input and generated datasets. The joint GAN with regularization yields a higher score, which confirms its improved support for rare or less frequent categories, which is vital for realistic agent-based modeling. 

\begin{figure*}[t]
    \centering
    \includegraphics[width=1\textwidth]{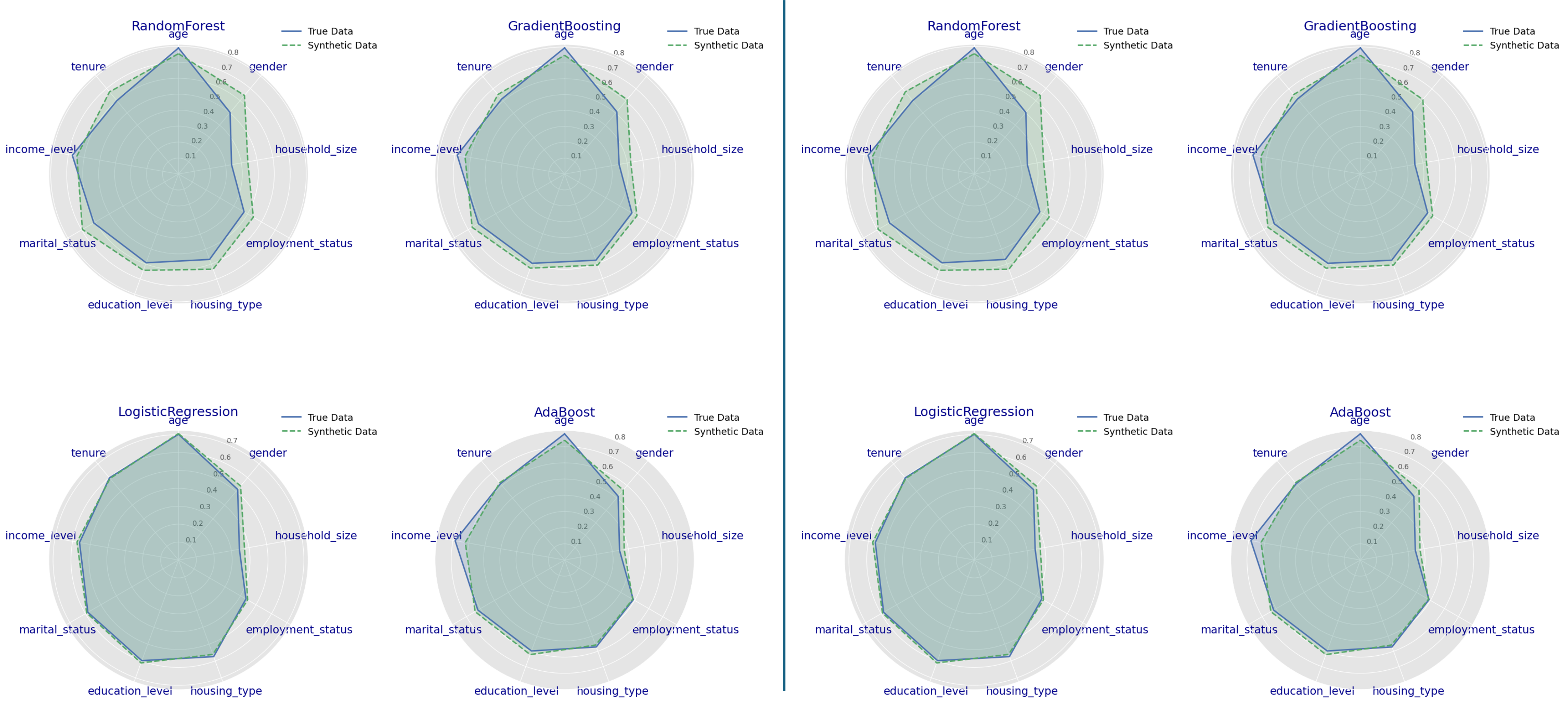} 
    \caption{\scriptsize Comparison of classification accuracies for individual attributes using real and synthetic data}
    \label{fig:ML efficiency}
\end{figure*}

Finally, the machine learning efficacy metric assesses whether synthetic data retain the predictive power of real data in downstream ML tasks. As shown in Figure ~\ref{fig:ML efficiency}, we train four classifiers (Logistic Regression, Random Forest, AdaBoost, Gradient Boosting) on synthetic data and test on real data. Based on Table ~\ref{tab:model_comparison} the first model achieves the lower relative error, substantially outperforming the Simple GAN. This demonstrates that the synthetic data generated by the proposed method generalize well and can substitute real data in predictive applications. 

Among all the metrics, the most significant improvement is observed in machine learning efficiency. This suggests that this metric should be considered the most important in evaluating synthetic data quality. The results confirm that both the joint fusion approach and the incorporation of the regularization term significantly improve the quality and applicability of the synthetic dataset in machine learning contexts. 

\begin{table*}[t]
\centering
\caption{Comparison of Models by Various Metrics}
\label{tab:model_comparison}
\resizebox{\textwidth}{!}{%
\begin{tabular}{l l ccc ccc ccc c cccc}
\hline
& & \multicolumn{3}{c}{\textbf{SRMSE}} & \multicolumn{3}{c}{\textbf{JSD}} & \multicolumn{3}{c}{\textbf{Correlation}} & \multicolumn{1}{c}{\textbf{PMSE}} & \multicolumn{4}{c}{\textbf{ML Efficiency}} \\
\textbf{Model} & \textbf{Data}
& \textbf{Marg} & \textbf{Bivar} & \textbf{Trivar}
& \textbf{Marg} & \textbf{Bivar} & \textbf{Trivar}
& \textbf{First} & \textbf{Second} & \textbf{Third}
& \textbf{LG} & \textbf{LG} & \textbf{RF} & \textbf{GB} & \textbf{AB} \\
\hline

\multirow{2}{*}{Joint WGAN + IGP}
& OD     & 0.09  & 0.18  & 0.28  & 0.007 & 0.021 & 0.040 & 0.036 & 0.042 & 0.061 & 0.12  & 0.086 & 0.076 & 0.072 & 0.087 \\
& Census & 0.12  & 0.27  & 0.43  & 0.014 & 0.049 & 0.083 & 0.05  & 0.063 & 0.072 & 0.162 & 0.032 & 0.144 & 0.096 & 0.047 \\

\multirow{2}{*}{Joint WGAN}
& OD     & 0.097 & 0.248 & 0.401 & 0.016 & 0.04  & 0.075 & 0.055 & 0.058 & 0.063 & 0.152 & 0.078 & 0.1   & 0.091 & 0.085 \\
& Census & 0.124 & 0.247 & 0.362 & 0.013 & 0.039 & 0.079 & 0.056 & 0.07  & 0.105 & 0.166 & 0.114 & 0.253 & 0.202 & 0.12 \\

\multirow{2}{*}{Simple WGAN}
& OD     & 0.112 & 0.267 & 0.42  & 0.019 & 0.042 & 0.081 & 0.052 & 0.061 & 0.069 & 0.168 & 0.096 & 0.145 & 0.18  & 0.098 \\
& Census & 0.132 & 0.332 & 0.392 & 0.017 & 0.052 & 0.088 & 0.058 & 0.082 & 0.099 & 0.186 & 0.198 & 0.37  & 0.269 & 0.18 \\
\hline
\end{tabular}%
}
\end{table*}

\subsection{Diversity and Feasibility}
Diversity in the generated data is closely related to recall, while feasibility is more aligned with precision. High recall indicates the model can reproduce a wide range of categories found in the real dataset and generate diverse outputs. In contrast, high precision shows that the categories generated by the model are valid and actually exist in the real data.

Recall is calculated as the ratio of True Positives (correctly generated categories that are present in the real dataset) to the sum of True Positives and False Negatives (categories that exist in the real dataset but are not generated by the model). Precision is calculated as the ratio of True Positives (correctly generated categories present in the real dataset) to the sum of True Positives and False Positives (categories that are generated by the model but do not exist in the real dataset). 

\subsection{Overall Quality}
We observed three notable patterns in the results presented in Table \ref{tab:three_final}. First, the Joint GAN model (0.869) outperforms the Simple GAN (0.846), which shows that simultaneous integration and generation from multiple sources yields better performance. Furthermore, incorporating a regularization term through the IGP further enhances the model. The Joint GAN\_IGP achieves the highest overall score of 0.881.

Second, while the overall improvement is evident, it is not uniformly distributed across all individual evaluation metrics. The most notable gains are observed in machine learning efficiency, where scores increase from 0.809 to 0.92. This indicates that traditional statistical metrics such as SRMSE, correlation, and distribution similarity may not fully capture model generalizability. Therefore, incorporating machine learning efficiency metrics can improve the reliability of assessing generalization.

Third, Although the difference between the best and worst final scores is only 3.5\%, the models vary meaningfully in performance and should not be considered equally effective. Apart from these metrics, the diversity and feasibility of the model outputs are also important. Therefore, it is important to complement quantitative evaluation with a closer look at whether the generated data is diverse, realistic, and structurally valid. 

The Joint GAN outperforms the simple GAN, with recall increasing by approximately 7\% and precision by about 15\%. This suggests that simultaneously generating and integrating data leads to greater model reliability compared to the sequential approach of first integrating and then generating. A regularization term is incorporated to enhance diversity while maintaining feasibility which is a challenging trade off, as novel feature combinations can result in infeasible samples. Nevertheless, recall improves by an additional 10\%, which shows that the model produces more diverse outputs. Precision also rises by around 1\%, a sign that feasibility improves. These results demonstrate that the Joint GAN effectively supports simultaneous integration and generation, and improves both the diversity and feasibility of the synthetic data.

\begin{table}[H]
\centering
\caption{Evaluation Score and Sampling/Structural Zeros Comparison}
\label{tab:three_final}
\resizebox{\columnwidth}{!}{%
\begin{tabular}{l c c c c c c | c c c}
\hline
\textbf{Model} & $S_{distance}$ & $S_{corr}$ & $S_{pmse}$ & $S_{cr}$ & $S_{ml}$ & \textbf{Final score} & \textbf{Recall} & \textbf{Precision} & \textbf{F1-score} \\
\hline
Joint GAN + IGP & \textbf{0.867} & \textbf{0.946} & \textbf{0.859} & \textbf{0.83} & \textbf{0.92} & \textbf{0.881} & \textbf{50.41\%} & \textbf{81.32\%} & \textbf{62.24\%} \\
Joint GAN       & 0.855 & 0.932 & 0.841 & 0.807 & 0.87 & 0.869 & 40.39\% & 80.92\% & 53.78\% \\
Simple GAN      & 0.837 & 0.930 & 0.823 & 0.818 & 0.809 & 0.846 & 33.45\% & 65.75\% & 44.34\% \\
\hline
\end{tabular}%
}
\end{table}

\section{Conclusion}

This study presents a framework for synthetic population generation by jointly integrating multi source datasets using a WGAN with gradient penalty and an IGP regularization term. The decision to fuse multiple datasets arises from the limitations of relying on a single data source, which often fails to capture the full complexity of individual behaviors and attributes. Census data offers comprehensive socio-demographic attributes but often lacks detailed behavioral patterns, whereas travel survey data captures rich mobility information but suffers from limited coverage and sample bias. By fusing these complementary datasets, our model learns a more comprehensive  representation of individual behavior, that captures both structural population characteristics and activity patterns. Unlike conventional methods that either use a single dataset or fuse data before generation, our approach enables simultaneous integration and synthesis, that preserves latent interdependencies between datasets. In addition to address data integration challenges, we also tackle a common problem in generative models known as mode collapse. To deal with this, we add a regularization term to our WGAN, which pushes the generator to create more diverse and realistic samples.

Our results demonstrate that the proposed Joint GAN outperforms both the simple GAN and the version without regularization in terms of universal evaluation metrics, feasibility, and diversity. Specifically, the inclusion of the IGP significantly enhances diversity by reducing mode collapse, while improving feasibility as reflected in higher recall and precision scores. The use of a universal evaluation metric also provides a more interpretable measure of synthetic data quality across multiple dimensions. While traditional statistical metrics such as SRMSE and correlation offer insight into distributional similarity, they may provide a misleading picture of model quality if used alone. Therefore, it is crucial to explicitly evaluate the diversity and feasibility of the generated data, as they reflect the model’s ability to capture rare but valid combinations and avoid unrealistic outputs.

By effectively addressing key challenges in population synthesis, this framework sets a foundation for generating high quality synthetic data suitable for use in ABM and other behaviorally simulation tools. The findings show the potential of advanced generative models not only to match statistical properties but also to preserve the diversity and feasibility of real world population characteristics.

A promising direction for future research involves extending the current framework to incorporate additional data sources beyond census and travel survey data. Integrating three or more heterogeneous datasets could significantly enhance the richness, diversity, and relevance of the synthetic population. Such multi dimensional integration enables more comprehensive agent based simulations that capture the complex relationships among sociodemographic, behavioral, and environmental factors. In addition, further contributions are needed in the design of regularization strategies to improve the diversity and feasibility in the generated outputs. While the IGP used in this study promotes diversity and feasibility, there is still a lack of diverse regularization techniques in the literature. exploring and developing new forms of regularization could further enhance both the diversity and feasibility of the generated data.

\section{Acknowledgments}
This study is funded by the Canada First Research Excellence Fund under the Bridging Divides program.

\bibliographystyle{IEEEtran}
\bibliography{references}

\newpage

\begin{IEEEbiography}[{\includegraphics[width=1in,height=1.25in,clip,keepaspectratio]{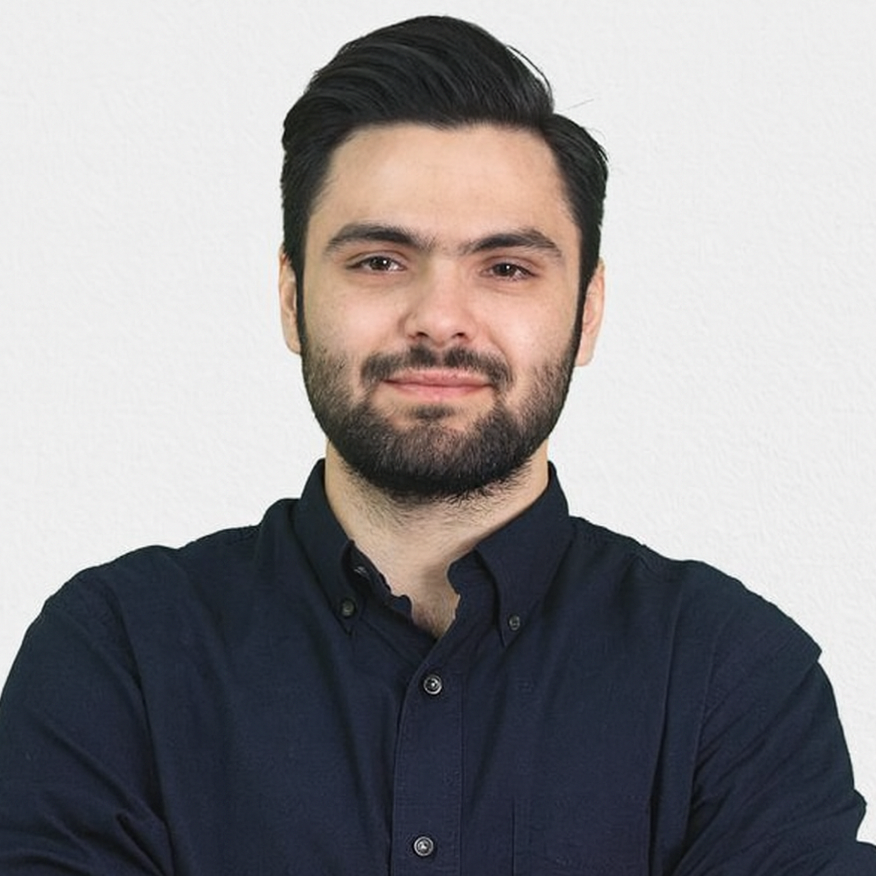}}]{Farbod Abbasi}
is currently a PhD student in the Department of Information and Systems Engineering at Concordia University in Montreal. His current research focuses on population synthesis and its applications in travel behavior analysis and transportation modeling.
\end{IEEEbiography}

\begin{IEEEbiography}[{\includegraphics[width=1in,height=1.25in,clip,keepaspectratio]{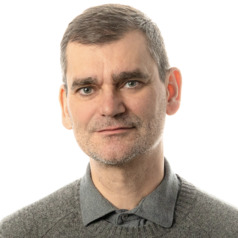}}]{Zachary Patterson}
is a Professor at the Institute of Information Systems Engineering
at Concordia University in Montreal, Canada. Zachary's research concentrates on:
the use of emerging technologies in transport-related data collection, computational
geospatial data processing, data analysis and inference with statistical and artificial intelligence, and discrete choice experiments. He is the Chair of the Transportation Research Board’s standing committee on Travel Data and Methods (AED17).
https://www.concordia.ca/faculty/zachary-patterson.html
\end{IEEEbiography}

\begin{IEEEbiography}[{\includegraphics[width=1in,height=1.25in,clip,keepaspectratio]{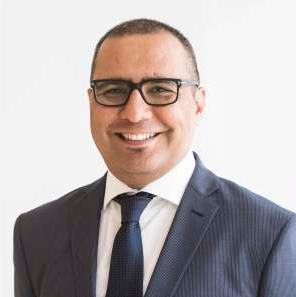}}]{Bilal Farooq}
received the Ph.D. degree from the University of Toronto, Canada, in 2011, the M.Sc. degree in computer science from the Lahore University of Management Sciences, Pakistan, in 2004, and the B.Eng. degree from the University of Engineering and Technology, Pakistan, in 2001. He is currently the Canada Research Chair of Disruptive Transportation Technologies and Services and an Associate Professor with Toronto Metropolitan University, Canada. His current research interests include the network and behavioral implications of emerging transportation technologies and services. He received the Early Researcher Award in Québec, in 2014, and Ontario, Canada, in 2018.
\end{IEEEbiography}

\vfill

\end{document}